\documentclass[11pt]{article}
\usepackage[hyperref]{ccl2022-en}
\usepackage{times}
\usepackage{algpseudocode, algorithm}
\usepackage{url}
\usepackage{makecell}
\usepackage{latexsym}
\usepackage{bm}
\usepackage{graphicx}
\usepackage{fancyhdr}
\usepackage{multibib}
\newcites{languageresource}{Language Resources}

\usepackage{soul}
\usepackage{ifthen}
\usepackage{booktabs}
\usepackage{bigstrut,multirow,rotating}
\usepackage{float}
\usepackage{stfloats}

\usepackage{epstopdf}
\usepackage{array}
\usepackage{CJKutf8}
\newcommand{\chinese}[1]{\begin{CJK}{UTF8}{gkai}{}#1\end{CJK}}
\usepackage[utf8]{inputenc}
\usepackage[T1]{fontenc}
\usepackage{xstring}

\renewcommand\footnotemark{}

\title{COMPILING: A Benchmark Dataset for Chinese Complexity Controllable Definition Generation}

\author{
	Jiaxin Yuan\textsuperscript{12*}\thanks{\textsuperscript{*}Equal contribution},
	Cunliang Kong\textsuperscript{123*},
	Chenhui Xie\textsuperscript{123}, \\
	{\bf \large 
		Liner Yang\textsuperscript{123\dag} \thanks{\textsuperscript{\dag}Corresponding author: Liner Yang (\href{mailto:lineryang@gmail.com}{lineryang@gmail.com})},
		Erhong Yang\textsuperscript{13}}
	\\ \textsuperscript{1}National Language Resources Monitoring and Research Center Print Media Language Branch, \\Beijing Language and Culture University 
	\\  \textsuperscript{2}School of Information Science, Beijing Language and Culture University 
	\\ \textsuperscript{3}Beijing Advanced Innovation Center for Language Resources, \\Beijing Language and Culture University 
	\\ 
	\texttt{jiaxinyuan625@gmail.com}
}

\date{}

\begin{document}
\maketitle
\begin{abstract}
  	The definition generation task aims to generate a word's definition within a specific context automatically. However, owing to the lack of datasets for different complexities, the definitions produced by models tend to keep the same complexity level. This paper proposes a novel task of generating definitions for a word with controllable complexity levels. Correspondingly, we introduce \textbf{COMPILING}, a dataset given detailed information about Chinese definitions, and each definition is labeled with its complexity levels. The COMPILING dataset includes 74,303 words and 106,882 definitions. To the best of our knowledge, it is the largest dataset of the Chinese definition generation task. We select various representative generation methods as baselines for this task and conduct evaluations, which illustrates that our dataset plays an outstanding role in assisting models in generating different complexity-level definitions. We believe that the COMPILING dataset will benefit further research in complexity controllable definition generation.
 
\end{abstract}

	\section{Introduction}
	
	Definition Generation (DG) is the task of describing the meaning that a word takes in a specific context.
	This task can help language learners by providing explanations for unfamiliar words.
	Recent researches \cite{Ishiwatari2019LearningTD,Zheng2021DecomposeFA} attempted to apply the task to the field of Intelligent Computer-Assisted Language Learning (ICALL), and have made a significant progress.
%	As a crucial technique in natural language processing, it is conducted to assist compilation of dictionary and plays a vital role in many other tasks such as word sense disambiguation.
%	Recent researches \cite{Ishiwatari2019LearningTD,Zheng2021DecomposeFA} attempt to apply this task to the field of Intelligent Computer-Assisted Language Learning (ICALL) to help language learners learn vocabulary by providing explanations for unfamiliar words and have made significant progress.

	Previous studies on DG mainly concentrate on generating different definitions for polysemous words \cite{Gadetsky2018ConditionalGO,Mickus-2019-mark,Reid-2020-vcdm}, or generating definitions with appropriate specificity \cite{Huang-2021-definition}.
	In these studies, researchers have faced various issues, such as the high complexity
	problem. High complexity definitions contain words that are more difficult than the defined word, and hence are labored for language learners to read and understand.
	Nevertheless, there have been few focuses on complexity controllable generation of definitions.
	A possible reason is that the complexities of definitions are not provided in currently existed datasets, which leads to the difficulty of automatic training and evaluation.
 
    Actually, the problems mentioned above are especially prominent in the language environment of Chinese. Definitions with suitable complexity are in urgent practical needs for Chinese as Foreign Language (CFL) learners.
	According to the Ministry of Education of China, by the end of 2020, more than 20 million foreign students are learning Chinese.
	But as \newcite{zhang-2011-duiwai} pointed out, since the difficulty of definitions is not considered, most existing dictionaries cannot meet CFL learner’s requirements.
	Besides, the existing Chinese learner dictionaries contain only a small number of words.
	For instance, the Contemporary Chinese Learner Dictionary (CCLD) only has about 6,400 words.
	In contrast, the Modern Chinese Dictionary (MCD), which is designed for native speakers, has about 69,000 words.

	Therefore, in this work, we focus on the task of generating definitions for CFL learners with appropriate complexities.
	At present, there are two datasets used for the Chinese definition generation task, but neither of them can meet the needs of this task.
	The most widely used CWN dataset \cite{Yang2020IncorporatingSI,Fan2020BERTChineseDM,Kong-2020-Toward} was built from the Chinese WordNet \cite{huang-2010-chinese}, which is a knowledge base of sense distinction\footnote{\url{http://lope.linguistics.ntu.edu.tw/cwn2}}.
	This dataset is limited in size with 8,221 words.
	\newcite{Zheng2021DecomposeFA} constructed a dataset from the 5th edition of MCD.
	But it only collects disyllabic nouns and verbs, and additional annotation of formation rules is required.
	Besides, both datasets didn't provide the complexity of definitions, which is essential information in the controllable generation.
	
	To enhance the study of this task, we propose to build a novel benchmark dataset named \textbf{COMPILING} (\textbf{C}hinese c\textbf{OMP}lex\textbf{I}ty contro\textbf{L}lable def\textbf{IN}ition \textbf{G}eneration).
	The dataset is large and of high quality, which contains 127,757 entries in total.
	Each entry consists of a word, an example, a definition, and two complexity measurements of this definition.
	More specifically,  we build the dataset by using two Chinese dictionaries, namely the CCLD and the 7th edition of MCD.
	The former collects fewer words, but the definitions are simpler.
	The latter is the opposite.
	By combining these two dictionaries, we obtain a large amount of definitions that vary in different complexities.
	
	In order to quantitatively measure the \textit{complexity} of definitions, we refer to the graded vocabularies formulated by HSK (Chinese Proficiency Test).
	HSK is set up to test the proficiency of non-native speakers.
	It has nine levels from easy to hard, and each level corresponds to a vocabulary.
	The COMPILING dataset contains an average level and a maximum level of each definition.
	
	We find that both dictionaries tend to use phrases rather than complete sentences as examples in some cases.
	For instance, the word “\chinese{规模}” (scale) has two example phrases in MCD (Modern Chinese Dictionary), which are “\chinese{规模宏大}” (large scale) and “\chinese{初具规模}” (begin to take shape).
	We think that short phrases might be helpful for language learners to understand, but complete sentences can provide more context in the automatic definition generation.
	Thus, we design an algorithm to expand the phrases into sentences (Section \ref{section:expand-algo}).
	
	We believe that this dataset can further enhance the research on Chinese complexity controllable definition generation, which could not only benefit the language learners, but also low literacy readers, as well as people with aphasia or dyslexia.
	We also provide baselines of mainstream generation methods as references (Section \ref{section:experiments}).
	
	In summary, our contributions are listed below:
	
	\begin{itemize}
		\item{We propose a novel task of generating definitions for a word with appropriate complexity. The task is of great use in helping CFL learners to learn the vocabulary.}
		\item{ We propose the \textbf{COMPILING} dataset that is of large scale and high quality. This dataset could serve as the benchmark of the task we proposed.}
		\item {We perform several experiments on the COMPILING dataset and the results demonstrate it could assist models to achieve effective complexity controllable generations.}
	\end{itemize}

	\section{ Related Work}
%\vspace{1ex}
    \subsection{Definition Generation}
	\newcite{Noraset2017DefinitionML} first proposed the definition modeling task and use word embeddings to generate definitions of the corresponding words.
	Referencing on the problem of word sense disambiguation, \newcite{Ishiwatari2019LearningTD} and \newcite{Gadetsky2018ConditionalGO} incorporated word contexts into definition modeling and demonstrated its effectiveness of distinguishing different meanings.
	Recent work \cite{Huang2021CDMCE} reformulates the task as generating descriptions using extracted knowledge.
	Research on Chinese definition modeling was first proposed by \newcite{Yang2020IncorporatingSI}, they adapted a transformer-based model and incorporated sememes into the model to provide more external semantic knowledge.
	\newcite{Fan2020BERTChineseDM} redefined the Chinese definition modeling as generating the corresponding definition for a target word and its context.
	\newcite{Zheng2021DecomposeFA} utilized the characteristics of Chinese by adding formation features to enhance definition modeling.
	Besides, there are also studies on multilingual definition generation \cite{Kong-2020-Toward} and combining extraction and generation for this task \cite{huang-2021-cdm}.
	
	Notably, \newcite{kong-etal-2022-multitasking} proposed to generate simple definitions employing a multitasking framework.
	Since the lack of a definition dataset with different complexities, they managed to generate both complex and simple definitions in an unsupervised way.
	
	Differently, we focus on building the benchmark dataset for different Chinese definition generation tasks and hope it could be beneficial for further research.
	
\subsection{Controllable Generation}
Controllable generation is widely adapted in kinds of language modeling tasks. For instance, data augmentation \cite{AminNejad2020ExploringTT}, dialog generation \cite{Firdaus2020EmoSenGS}, storytelling \cite{GoldfarbTarrant2020ContentPF}, and so on. And the objects controlled in different studies vary from each other. Specifically, considering the significance of sentiment in poetry definition, \newcite{Chen2019SentimentControllableCP} proposed a model to generate poetry with controllable emotions. \newcite{Gao2019DifficultyCG} first presented a framework to develop questions about specific answers that meet target difficulty levels. To attract more readers, \newcite{Jin2020HooksIT} introduced a headline generation model to produce enticing titles with target three styles. Likewise, in order to explore and release the practical value of definition generation, we propose the complexity controllable definition generation task committed to producing definitions satisfying users of all levels.

Currently, the most controllable generation tasks are achieved based on pre-trained learning models. And \newcite{Zhang2022ASO} summarized the common methods as Finetuning, Retrain PLMs, and Post-Process. And we utilize the first method to control the complexity of the definition more efficiently.

	\subsection{Prompt Learning}
	In recent years, the pre-trained model with fine-tuning has gradually become the mainstream of natural language processing tasks. Due to the complex training objectives and large hyperparameter groups, large-scale pre-training models can effectively extract features from a large amount of supervised and unsupervised data. By storing the learned knowledge in parameters and fine-tuning the model for specific tasks, the same model can be applied to a series of downstream natural language processing tasks \cite{Han2021PreTrainedMP}.
	
	Prompt learning is a method of fully learning knowledge by adding additional text to the model’s input. Prompt can be divided into artificial and automatic construction according to the text attached to the input \cite{Han2021PreTrainedMP}. Among them, automatically constructed prompts are divided into discrete and continuous ones. A discrete prompt refers to the fact that the constructed prompt is composed of actual text symbols, and applicable tasks include text classification \cite{Han2021PTRPT}, text generation \cite{Zheng2021ExploringPF}, etc. 
	
	Although the combination of pre-training and fine-tuning methods can be adapted to most NLP tasks, when it comes to each specific task, the number of parameters that need to be adjusted for are vast. By adopting prompt learning, the pre-training model can be applied to the required tasks by only modifying the part of the prompt for different downstream tasks. Therefore, the training process will become more efficient.
	
	\section{Problem Formulation}
In this work, we aim to generate a definition $\bm d^c$ with appropriate complexity $c$, for a given word and example sentence $(w^*, \bm e)$.
This task is feasible because the word and it's corresponding definition should be assumed to have the same semantics.
A common solution is to predict tokens in the definition one by one, depending on the previous words and the other conditions, which can be formulated as:
\begin{equation}
P(\bm d^c | w^*, \bm e, c) = \prod_{t=1}^T P(\bm d^c_t | \bm d^c_{<t}, w^*, \bm e, c),
\end{equation}
where $d^c_t$ is the $t$-th token in the definition, and $T$ is the total length of definition. 
Each probability distribution can be approximated by the following equation:
\begin{equation}
P(\bm d^c_t | \bm d^c_{<t}, w^*, \bm e, c) \propto \exp(Wh_t/\tau),
\end{equation}
where $W$ is a matrix collecting word vectors, $h_t$ is a vector summarizing inputs at current time-step, and $\tau$ is a hyper-parameter for temperature, set to 1 in default.

\section{Dataset Construction}
The source corpora are extracted from the MCD and CCLD, both published by the Commercial Press.
For corpus from MCD and CCLD, we process them separately with the same construction methods and finally put them together.

The construction of the COMPILING dataset is divided into three stages: data structured annotation, example sentences expansion, and post-processing. First, we propose a strategy for building structured datasets due to the high complexity and compact construction of automatically extracted data. In this phase, we set up a platform. It not only helps annotators proofread and audit corpus data more efficiently but is also conducive for us to check and collect data. Besides, since the context of a targeted word in the dictionary is always a collocation instead of a complete sentence, we then conduct expanding context to enhance the overall abundance of language for our proposed datasets. Furthermore, to divide definitions into different complexity levels, we calculate the HSK level of each description.

\subsection{Data Structured Annotation}
%In the beginning, we adapt Optical Character Recognition (OCR) technique to scan dictionaries. The initial data produced by OCR is disorganized and complex in structure, which is problematic to conduct automatic processing.

In the beginning, we collect initial data and find they are
disorganized and complex in structure, which is problematic to conduct automatic processing. Hence, we start up data structured annotation. 
To better manage and boost the whole process, we build up a platform before the formal annotation and deploy it on two servers, one for corpus from MCD, and the other for corpus from CCLD. This platform could not only serve specifically for this task, but it is also appropriate for the construction of any resource by replacing the data.

Concentrating on tackling the problem of disorganized data, we suggest a series of rules for annotation. For a particular word, its attached contents include its spell, definition, example sentences of the usage of a specific definition, and so on. Hence, we propose to add labels before corresponding contents to distinguish different types of data, which is conducive for computers to extract this information based on their labels automatically. Both dictionaries have instructions illustrating the meta-information, such as the organization of entries, the style of definitions and examples, and basic usages. We invite a student who majors in linguistics to formulate the annotation guidelines based on the instructions, which will be the reference for annotators. By doing so, we hope annotators could restore that language information and the relationships between them to a large extent. Then, we invite 20  students majoring in linguistics to annotate the corpora on our platform regarding the guidelines. This phase lasted for two months.

	\begin{algorithm}[htb]
	\renewcommand{\algorithmicrequire}{\textbf{Input:}}
	\renewcommand{\algorithmicensure}{\textbf{Output:}}
	\caption{Example Sentences Expansion}
	\label{alg:1}
	\begin{algorithmic}[1]
		\Require phrase $p$, corpus $C$
		\Ensure examples $E$
%		\STATE $topN \gets 5$;
		\State $D \gets \{\}, E \gets []$

%		\STATE $final \gets$ new Hashtable;
		\For{ $sentence$ \textbf{in} $C$ }
		\If {$p$ \textbf{in} $sentence$}
			\State $score \gets pplScore(sentence)$
			\Comment Compute the PPL score for each sentence.
%			\State add $\{sentence: score\}$ into $D$;
			\State $D[sentence] \gets score$
		\EndIf 
		\EndFor
		
		\State $sortedExamples \gets descSortByValue(D)$
		\Comment Descendant sort by the scores.
%		\STATE $sortedContexts \gets sorted(dict.keys(),key=lambda x:x[1],reverse=True)$
%		\STATE $$;
%		\FOR{ k in top $n$ }
%		\STATE insert contexts[k] into final;
%		\ENDFOR
		\For {$i=0 \to topN$}
		\Comment $topN$ is set to 5 in practice.
%		\State insert $sortedExamples[n]$ into $T$;
		\State $E.add(sortedExamples[i])$
		\EndFor
%		\STATE $S$ $\leftarrow$ final;
%		\STATE \textbf{return} $S$;
	\end{algorithmic}  
\end{algorithm}\label{contextexpansion}

\subsection{Example Sentences Expansion} \label{section:expand-algo}
While the information extracted from dictionaries is large and abundant, the context attached to the targeted words given in dictionaries is too short to provide enough knowledge for the model to learn and generate descriptions. In the second stage of construction, considering the significance of sentences, we start up example sentence expansion. For contexts without sufficient length in the original corpus, we tend to find sentences with a longer length and higher quality in the new canon for replacement, and the specific process is as follows.
We first screened each example sentence in the annotated texts. We set the length threshold to six, and if the length of the initial context is longer than the threshold, we will retain the sentences; otherwise, we will find longer sentences with more abundant information in the new corpus to cover the original ones. It is worth noting that if a term contains more than one sentence (collocation), for each sentence (collocation), we will replace it with new matching contexts.

We design Algorithm \ref{alg:1} to match and gain new high-quality sentences. Given the ambiguity of most words, we utilize an allocation as the input of  Algorithm \ref{alg:1} instead of a phrase to ensure the found sentences contain the corresponding usage of a specific definition. As shown in Algorithm \ref{alg:1}, we collect all the sentences that fit the requirements and grade them by utilizing Perplexity (PPL)\footnote{\url{https://huggingface.co/docs/transformers/perplexity}}, which is one of the most common metrics for evaluating language fluency. Eventually, the top five sentences in the rankings are designated to replace those original short contexts.

\subsection{Post Processing}
\label{hsk}
\paragraph{Difficulty classification}
The most crucial step of constructing a complexity-controlled dataset is integrating the difficulty level of definition into the dataset. We utilize the HSK metric to represent the complexity degree. HSK\footnote{\url{http://www.chinesetest.cn}}, called the Chinese Proficiency Test, set to evaluate the Chinese proficiency and application of non-native speakers. It is divided into nine levels, and the difficulty increases progressively from low to high. For convenience, we regard the seventh, eighth, and ninth levels as a whole. Finally, we set seven complexity levels of HSK, and each level corresponds to a vocabulary. For words that are not included in the first seven-level, we classify them as the highest level.
\paragraph{Entry construction}
Besides, For each definition, we first conduct word segmentation, then calculate the average and highest HSK level, and combine the HSK level into the dataset. Eventually, each entry of the COMPILING dataset consists of a target word, its definition, the average and highest HSK level, and the contexts of the corresponding usage of this description.

	\section{Dataset Analysis}\label{dataset}

\begin{table}[htbp]
	\centering
	\setlength{\tabcolsep}{12pt}
	\caption{The main statistics of the COMPILING dataset.}
	
	\begin{tabular}{lcrrcrr}
		\toprule[1pt]
		\multirow{2}{*}{Datasets} & & \multicolumn{2}{c}{Count} & & \multicolumn{2}{c}{Average Length} \\
		\cmidrule{3-4} \cmidrule{6-7}
		 & & Words & Entries & & Definition & Context \\
		\midrule
		MCD   & & 67,801 & 101,314 & & 13.8  & 27.5  \\
		CCLD  & & 6,502  & 26,443 & & 13.4  & 20.4 \\
		\bottomrule[1pt]
	\end{tabular}%
	\label{tableone}%
\end{table}%
As mentioned before, the smallest unit of the COMPILING dataset consists of five parts. In particular, if a word is polysemous or has numerous contexts, they are regarded as distinct entries. For instance, as shown in Table \ref{addlabe1}, the word
 “\chinese{收拾}” (clear up) has four different definitions, and each of them follows an example sentence. Hence there are four entries of  “\chinese{收拾}” (clear up) in total. 

As shown in Table \ref{tableone}, we analyze statistics of data extracted from MCD and CCLD, respectively. Table \ref{tabletwo} shows the basic statistics of the COMPILING dataset and another dataset of Chinese definition modeling. For training, the given definitions of each entry are seen as the ground truth. 

\begin{table}[htbp]
	\centering
	\caption{Example entries of COMPILING dataset.}
	\resizebox{\textwidth}{!}{
	\begin{tabular}{cccccc}
		\toprule
		Word & Definition & Average & Maximum & Sentence & Source \\
		\midrule
		\makecell[c]{\chinese{收拾}\\clear up} & \makecell[c]{\chinese{使变干净整齐；整理} \\To make clean and tidy} & 2     & 3     & \makecell[c]{\chinese{东西都收拾好了，可以出门了。}\\ With everything packed up, \\we're ready to go.} & CCLD \\
		\midrule
		\makecell[c]{\chinese{收拾}\\repair} & \makecell[c]{\chinese{使有毛病的东西功能正常；修理}\\To make something\\ defective function properly} & 2     & 4     & 
		\makecell[c]{\chinese{我的手机坏了，得找厂家收拾一下。}\\My phone is out of order so I have to ask\\ manufacturer for help.} & CCLD \\
		\midrule
		\makecell[c]{\chinese{收拾} \\settle} & \makecell[c]{\chinese{整理；整顿} \\Put in order} & 4     & 6     & \makecell[c]{\chinese{冬储夏衣，夏藏冬衣，收拾屋子，还要照看外孙女。}\\Store summer clothes in the winter, hide\\ winter clothes in the summer, clean \\the house, and look after her granddaughter.} & MCD \\
		\midrule
		\makecell[c]{\chinese{收拾}\\kill} & \makecell[c]{\chinese{消灭；杀死} \\Eliminate} & 8     & 10    &  \makecell[c]{\chinese{据点的敌人，全叫我们收拾了。} \\All the enemies in the stronghold \\have been eliminated.} & MCD \\
		\bottomrule
	\end{tabular}}%
	\label{addlabe1}%
	\vspace{2.0em}
\end{table}%

\begin{figure}[ht]
	\centering
	\includegraphics[width=.8\textwidth]{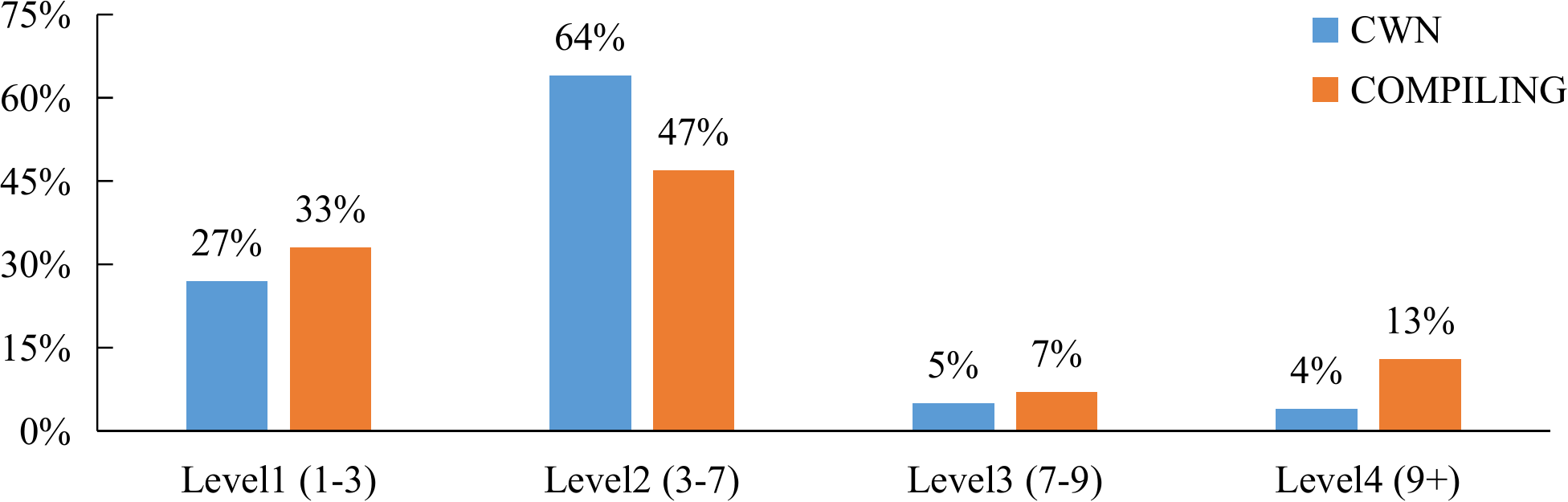}
	\caption{The distribution of average HSK level in CWN and COMPILING.}
	\label{distribution}
\end{figure}

\begin{table}[!htbp]
	\vspace*{-1cm}
	\centering
	\setlength{\tabcolsep}{12pt}
	\caption{Statistics of Chinese definition modeling datasets.}
	\footnotesize
	\begin{tabular}{lrrrrrr}
		\toprule[1pt]
		\multirow{2}{*}{Datasets} & & \multicolumn{2}{c}{Count} & & \multicolumn{2}{c}{Average Length} \\
		\cmidrule{3-4}
		\cmidrule{6-7}
		 & & Words & Entries & & Definition & Context \\
		\midrule
		CWN   & & 8,221 & 84,542 & & 9.07  & 21.57  \\
		\textbf{COMPILING} & & 74,303 & 127,757 & & 13.60  & 23.95  \\
		\bottomrule[1pt]
	\end{tabular}%
    	\label{tabletwo}%
\end{table}%
To better highlight the complexity degree of the dataset, we set levels 1-3 in HSK as the simple grade, levels 3-7 as the medium grade, and levels 7-9 and 9+ as hard quality. We count the HSK level distribution of definitions of COMPILING, as shown in Figure \ref{distribution}.

%\begin{figure}[ht]
%	\centering
%	\includegraphics[width=.7\textwidth]{imgs/fenbu.jpg}
%	\caption{The distribution of (Average) HSK level in CWN and COMPILING.} 
%	\label{distribution}
%\end{figure}

The distribution of definitions in the COMPILING dataset in the three levels is closer than CWN. Given the particularity of the Complexity Controllable definition generation task, it is necessary to construct a dataset including entries covering all difficulty levels. In this way, the model can learn and distinguish the complexity of descriptions, hence generating a new definition of a word with a target complexity level.

Hence, the COMPILING dataset could be applied to both general definition generation tasks and those which incorporate the complexity of definitions, demonstrating its value in being as a benchmark dataset.

	\section{Experiments}\label{section:experiments}
	\subsection{Baselines}\label{model}
	This section introduces several methods for common generation tasks, which can serve as baselines for our proposed task.
	
	\paragraph{LOG-CaD}
	LOG-CaD \cite{Ishiwatari2019LearningTD} is a model for generating descriptions for words and phrases.
	This model summarizes clues from the static, contextualized, as well as character-level embeddings of the given word, and then employs an LSTM-decoder for the generation.
	A gated attention mechanism is employed to capture and filter information from the embeddings during decoding.
	
	\paragraph{Transformser}
	We treat the task as a special type of single language translation and directly use the original transformer model proposed by \newcite{Vaswani-2017-attention}.
	We concatenate the word and example sentence as the input sequence and train the model to generate the definition.
	We use the same approach to deal with the input and output in BERT and BART models.
	All hyper-parameters are set according to the original paper for a fair comparison.
	
	\paragraph{BERT}
	Pretrained language models have been widely used in various NLP tasks in recent years.
	By obtaining prior knowledge during pretraining, the PLMs can encode the input sentence more effectively.
	Thus, we use the Chinese-bert-base \cite{Devlin-2019-bert} model to initialize all the parameters in a transformer encoder and employ a transformer decoder for the generation.
	Note that the decoder is trained from scratch without initialization.
	
    \paragraph{BART}
    Unlike BERT, BART \cite{Lewis-2019-bart} is a pretrained encoder-decoder language model, which is more suitable for generation tasks.   
    Since the monolingual BART only support English, we use the multilingual version of BART and set both source and target language as Chinese for this task. 
    
     \paragraph{MT5}
    T5 is one of the representative pre-training language models. It considers all NLP tasks as a uniform text-to-text paradigm. mT5 \cite{Xue2021mT5AM} is a multi-language variant of T5, and its performance on various benchmark tasks is generally outstanding. Therefore, we choose mT5 to perform the prompt learning method.
    
    \begin{table}[h]
    	\centering
    	\setlength{\tabcolsep}{15pt}
    	\caption{Datasets divided by HSK level.}
    	\begin{tabular}{lcc}
    		\toprule[1pt]
    		Complexity & HSK   & Entries \\
    		\midrule
    		Easy & 1-3   & 48,458 \\
    		Medium & 4-7   & 53,945 \\
    		Hard & 7+    & 25,354 \\
    		\bottomrule[1pt]
    	\end{tabular}%
    	\label{tablefour}%
    \end{table}%
	
	\subsection{Settings}\label{sett}
%	\paragraph{Definition Generation}
	As a benchmark dataset introduced to enhance the Chinese definition generation task, we set up the experiments to verify the effectiveness of the COMPILING dataset. 
%	We divide the dataset into training, validation, and test sets according to 8:1:1. The training data are fine-tuned according to the input formats of different models.
	
	\paragraph{Regardless of complexity levels}
	We first design the experiment to evaluate the overall performance of the baseline models on our dataset.
	In this setting, we train the models using the entire training set, despite of the different complexity levels.
	And the purpose of this setting is to provide a comparison standard for other experiments.
	We divide the dataset into training, development, and test sets according to 8:1:1. The training data are fine-tuned according to the input formats of different models.
	
	\paragraph{Complexity specific models}
	To evaluate the significant role of the COMPILING dataset in generating definitions across various difficulties, we set up an experiment to train the model on different complexity-level sub-datasets.
	First of all, we split the dataset into three subsets on basis of the average HSK level. As shown in Table \ref{tablefour}, the HSK levels of definitions in Easy Set are between 1 to 3, Medium Set corresponding to level 4-6, and Hard Set corresponding to level 7+.
	Then we split each subset into training, development, and test sets according to the ratio of 8:1:1.
	Finally, we fine-tune the BART model utilizing these three training sets, and hence getting three models. Each one could generate definitions with its corresponding complexity level. 

    \paragraph{Unified model based on prompt learning}
    To assist the model to generate descriptions with different complexity of demand, we adopt the method of prompt learning. It allows the model to learn by adding tokens that represent difficulty information to the inputs, such as <extra\_id\_1> for level 1 (lowest), <extra\_id\_2> for level 2, and so on. The training set is formed by prefacing each definition of the COMPILING dataset with the corresponding special tokens. Each entry of the final dataset includes: <extra\_id\_x>, target word, its corresponding definitions and context.
    During the training phase, the model encodes both complexity and definition information. In the analysis stage, aiming to verify the effectiveness of this method, we select 10 entries from the test set of the COMPILING dataset. For each entry, only its difficulty token is modified with the other information keep remaining, so as to construct two copies of the entry. 
    It is worth noting that the principle of constructing the new complexity tokens is, that the two new entries and the original one(a group of data) differ by at least 2 levels or more, which means they can represent easy, medium, and hard complexity respectively. For example, if the definition of the source entry is specified with the difficulty as 3, the complexities of the two copies of it need to be constructed as at least 1 and 5.
    Finally, a total of 30 entries are included in the new test set. Then, we perform the model on this new test set to observe whether the generated definitions are differentiated in line with their specified complexity.
    
	\subsection{Evaluation Metrics} 
	
	In order to better analyze and quantify the experimental results, we select three evaluation metrics: BLEU \cite{Papineni2002BleuAM}, NIST \cite{Doddington2002AutomaticEO} and HSK, which are used to comprehensively evaluate the quality and complexity level of generated definitions. 
	
    \paragraph{BLEU}
    BLEU (Bilingual Evaluation Understudy) \cite{Papineni2002BleuAM} was originally proposed for the evaluation of machine translation research. The core of BLEU is to separately calculate the N-gram in the generated and the reference sentence, and then compare them one by one to count the times that can be matched. The higher times illustrate higher accuracy. However, the shorter reference segment always leads to more co-occurrence times, which means the shorter generated definitions tend to get a higher BLEU score.
 
 	\paragraph{NIST}
 	On the basis of BLEU, NIST (National Institute of Standards and Technology) \cite{Doddington2002AutomaticEO} adds the calculation of the information weight of N-gram. 
 	While the BLEU simply sums up the number of N-grams, the NIST sums up the information weights and then divides it by the number of N-gram segments in the whole sentence. In this way, the weightage of those N-grams which appear less frequently will be heavier.
 	
 	\paragraph{HSK}
 	As mentioned in section \ref{hsk}, HSK is a test set to evaluate the Chinese proficiency and application ability of non-native Chinese speakers. Based on the purpose of assisting CFL learners to understand Chinese well, we select HSK to measure the complexity level of definitions. Besides, we set seven difficulty levels (scores) of HSK and each of them corresponds to a vocabulary. The final level of a definition is determined by the average score of its segments.
	\subsection{Results and Analysis} 
	\label{performance_on_model}
	\paragraph{Regardless of complexity levels}
	% Table generated by Excel2LaTeX from sheet 'Experiments-newz
	
	We report the experimental results on the entire COMPILING dataset in Table \ref{olala}.
	The results show that PLMs outperforms the other two methods in terms of the BLEU and NIST scores apparently.
	However, the results of BERT and BART models diverged on these two metrics.
	Since NIST assigns different weights to tokens, we believe it better reflects the model’s performance.
	We confirmed this by reading the generated samples. 
		\begin{table}[htbp]
		\centering
		\caption{Experiment results on the COMPILING dataset.}
		\begin{tabular}{lccccccc}
			\toprule
			\multirow{2}{*}{ Models} & \multicolumn{3}{c}{Dev} &       & \multicolumn{3}{c}{Test} \\
			\cmidrule{2-4} \cmidrule{6-8}
			 & BLEU  & NIST  & HSK   &       & BLEU  & NIST  & HSK \\
				\midrule
			LOG-CaD & 27.66  & 25.55  & 3.74  &       & 27.71 & 27.88 & 3.85 \\
			Transformer & 28.61 & 25.85 & 3.92  &       & 28.58  & 31.00  & 3.96 \\
			BERT  & \textbf{32.95 } & 29.66  & 4.05  &       & \textbf{32.03} & 30.56 & 4.08 \\
			BART  & 29.49  & \textbf{36.90 } & \textbf{4.76}  &       & 30.63 & \textbf{42.79} & \textbf{4.80}  \\
			\bottomrule
		\end{tabular}%
		\label{olala}%
	\end{table}%
	We also notice that as the model performance improves, so does the average HSK level of the generated definitions.
	This phenomenon is because simpler words are used more frequently, and hence are more easily learned by models.
	As the modeling ability improves, the better-performing models learn to use more complex words.
	This can be challenging for future complexity controllable definition generation works, i.e., improving the	performance and reducing the generation complexity at the same time.
	
	\paragraph{Complexity specific models}
	Table \ref{multi} illustrates experiment results on three different subsets.
	As listed in the table, we not only test on the subset in which the model is trained, but also on other subsets.
	Generally, all the models perform best on the subset it was trained, and poorly on other subsets.
	Moreover, the performance decays as the complexity level between the model and data increases.
    Definitions with different complexity have different lexical and syntax, resulting in poor cross-complexity generalization.
    Besides, we found that even on different test sets, definitions generated by the same model have similar complexity.
    
    	\begin{table}[hbp]
    	\centering
    	\caption{Experiment results in terms of complexity controllable generation on three test sets.}
    	\begin{tabular}{lccccccccccc}
    		\toprule
    		\multirow{2}{*}{Models} & \multicolumn{3}{c}{Easy Set} &       & \multicolumn{3}{c}{Medium Set} &       & \multicolumn{3}{c}{Hard Set} \\
    		\cmidrule{2-4}\cmidrule{6-8}\cmidrule{10-12} & BLEU  & NIST  & HSK   &       & BLEU  & NIST  & HSK   &       & BLEU  & NIST  & HSK \\
    		\midrule
    		BART-Easy & \textbf{32.44 } & \textbf{64.40 } & 2.40  &       & 21.56  & 27.61  & 2.73  &       & 25.89  & 7.95  & 2.74  \\
    		BART-Medium & 22.92  & 24.59  & 4.70  &       & \textbf{27.69 } & \textbf{40.68 } & 4.86  &       & 29.37  & 16.09  & 5.01  \\
    		BART-Hard & 22.49  & 3.55  & \textbf{8.46}  &       & 23.70  & 7.04  & \textbf{8.45}  &       & \textbf{46.57 } & \textbf{18.22 } & \textbf{8.76}  \\
    		\bottomrule
    	\end{tabular}%
    	\label{multi}%
    \end{table}%
    
\begin{figure*}[htbp]
		\centering
		\includegraphics[width=0.85\textwidth]{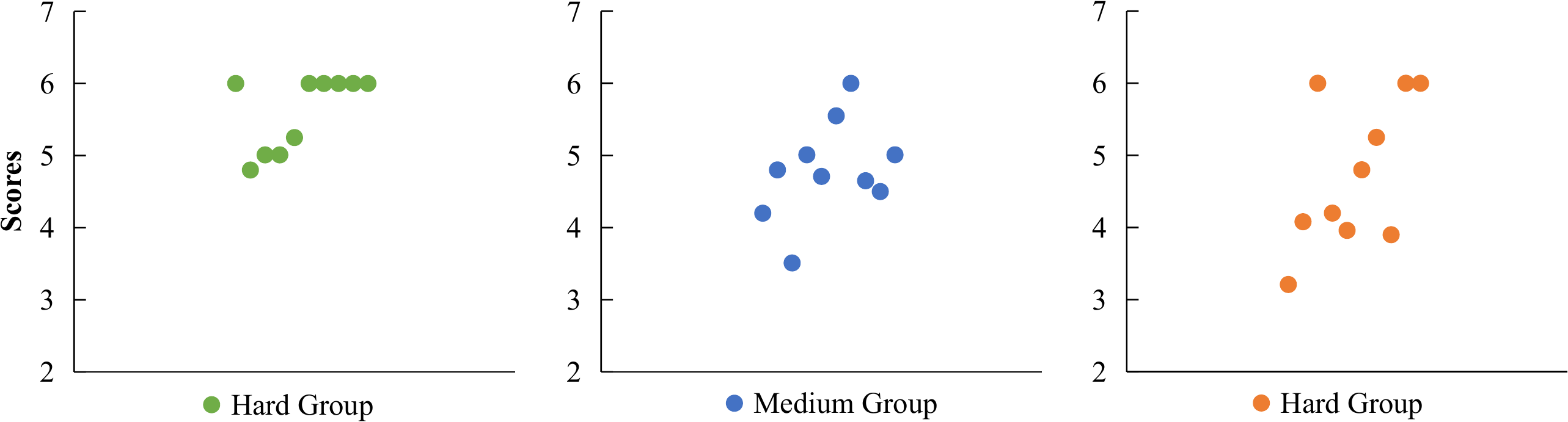}
		\caption{The automatic evaluation results. For example, the scatters of the Hard Group represent those definitions that are specified as the hardest, and the ordinate corresponds to the scores obtained by the automatic rating.} 
		\label{scatter}
	\end{figure*}

	\paragraph{Unified model based on prompt learning}
	MT5-base \cite{Xue2021mT5AM} was selected as the benchmark model in this experiment. The best PPL obtained from the definitions generated on the validation set is 38.44. The BLEU and NIST of the model on the test set are 27.42 and 4.66, respectively.
	The model generates interpretations based on the new test mentioned in Section \ref{sett}. Table 7 lists two examples where it is fairly obvious that the resulting definitions are differentiated and conform to the expectations for their specified complexity levels.
% Table generated by Excel2LaTeX from sheet 'Sheet2'
% Table generated by Excel2LaTeX from sheet 'Sheet2'
% Table generated by Excel2LaTeX from sheet 'Sheet2'a
	To evaluate the complexity of generating definitions more accurately, we adopt automatic evaluation, ranking the difficulty of each group\footnote{Each group of data refers to one original entry and its two copies, their specified complexity of definition is different and other information keep the same.}.
	The automatic evaluation is based on the Chinese Text Complexity Analysis Platform (CTAP)\footnote{\url{http://ctap.wenmind.net}} \cite{cui-2022-ctap}. We selected the features of word diversity and word density that reflect the difficulty of paraphrases and calculated the scores of definitions in each group based on the above features. Finally, the scatter distribution diagram is shown in Figure \ref{scatter}.
	 It can be seen that the complexity score of the Hard Group is mainly above 5, and the number of definitions with the highest score is the largest. The definition in the Easy Group scored the lowest overall score. This means the difficulty level of the model-generated interpretations obtained by automatic evaluation is roughly in line with expectations.
	The result proves the effectiveness of prompt learning on complexity controllable task, but since the difference in the overall distribution of scattered points in each group in the figure is not particularly obvious, it also reflects that there is room for exploration and improvement of this task in the future.

		\section{Conclusion}
In this work, we propose a novel task of generating
Chinese complexity controllable definitions for a given word and example sentence. This task is of great use in
helping CFL learners and low literacy readers. Meanwhile,
we introduce the COMPILING dataset, which
is a benchmark adapting to kinds of definition generation tasks. We also provide several baselines for this task, among which the prompt learning method better
assist models in generating definitions with specified complexity. Nevertheless, the experimental results also show that this task is challenging, and the performance needs further improvement.

	\section*{Acknowledgments}
	This work was supported by the funds of Research Project of the National Language Commission (No. ZDI135-131, No. ZDI145-24) and Fundamental Research Funds for the Central Universities in BLCU (No. 21PT04). We would like to thank all anonymous reviewers for their valuable comments and suggestions on this work.

%\bibliographystyle{ccl}
%\bibliography{ccl2022-en}

\bibliographystyle{ccl}
\bibliography{ccl2022-en}

\end{document}